\documentclass[10pt]{article} % For LaTeX2e
\usepackage[preprint]{tmlr}
\usepackage{amsmath,amsfonts,bm}

\def\eqref#1{equation~\ref{#1}}
\def\1{\bm{1}}

\DeclareMathAlphabet{\mathsfit}{\encodingdefault}{\sfdefault}{m}{sl}
\SetMathAlphabet{\mathsfit}{bold}{\encodingdefault}{\sfdefault}{bx}{n}

\usepackage{hyperref}
\usepackage{url}

\usepackage{multirow}
\usepackage{array}

\usepackage{url}

\usepackage{graphicx}

\usepackage{smartdiagram}
\newtheorem{Remark}{Remark}

\title{MAC: A unified framework boosting low resource automatic speech recognition}

\author{\name Zeping Min \email zpm@pku.edu.cn \\
      \addr School of Mathematical Sciences\\
      Peking University
      \AND
      \name Qian Ge \email geqian@pku.edu.cn \\
      \addr Academy for Advanced Interdisciplinary Studies \\
      Peking University
      \AND
      \name Zhong Li \email lzhong@microsoft\\
      \addr Microsoft Research Asia 
      \AND
      \name Weinan E \email weinan@math.pku.edu.cn\\
      \addr Center for Machine Learning Research \\
      Peking University
      }

\def\month{MM}  % Insert correct month for camera-ready version
\def\year{YYYY} % Insert correct year for camera-ready version
\def\openreview{\url{https://openreview.net/forum?id=XXXX}} % Insert correct link to OpenReview for camera-ready version

\begin{document}

\maketitle

\begin{abstract}
We propose a unified framework for low resource automatic speech recognition tasks named meta audio concatenation (MAC).
It is easy to implement and can be carried out in extremely low resource environments.
Mathematically, we give a clear description of MAC framework from the perspective of bayesian sampling.
In this framework, we leverage a novel concatenative synthesis text-to-speech system  to boost the low resource ASR task. 
By the concatenative synthesis text-to-speech system, we can integrate language pronunciation rules and adjust the TTS process. 
Furthermore, we propose a broad notion of meta audio set to meet the modeling needs of different languages and different scenes when using the system.
Extensive experiments have demonstrated the great effectiveness of MAC on low resource ASR tasks. For CTC greedy search, CTC prefix, attention, and attention rescoring decode mode  in Cantonese ASR task, Taiwanese ASR task, and Japanese ASR task the MAC method can reduce the CER by more than 15\%.
Furthermore, in the ASR task, MAC beats wav2vec2 (with fine-tuning) on common voice datasets of Cantonese  and gets  really competitive results on common voice datasets of Taiwanese and Japanese. Among them, it is worth mentioning that we achieve a \textbf{10.9\%} character error rate (CER) on the common voice Cantonese ASR task, bringing  about \textbf{30\%} relative improvement compared to the wav2vec2 (with fine-tuning).
\end{abstract}

\section{Introduction}
Automatic speech recognition (ASR) is a traditional task with wide application. Before deep learning became popular, HMM-GMM \citep{rodriguez1997speech} models, which have an elegant mathematical form, were widely used for ASR tasks. It performed well in some simpler speech recognition scenarios. And the most famous HMM-GMM speech recognition toolkit was Kaldi \citep{povey2011kaldi}. As the rise of deep learning, a large number of end-to-end speech recognition models have emerged, such as Speech-transformer \citep{dong2018speech}, Conformer \citep{gulati2020conformer}, Espnet \citep{watanabe2018espnet}, Wenet \citep{yao2021wenet}, LAS \citep{chan2015listen}, etc. Furthermore, there are also many pretrained large models, such as vq-wav2vec \citep{baevski2019vq}, wav2vec \citep{schneider2019wav2vec} and wav2vec2 \citep{baevski2020wav2vec} \citep{conneau2020unsupervised}.

However, whether for a HMM-GMM \citep{rodriguez1997speech} model or advanced end-to-end models such as wenet \citep{yao2021wenet}, obtaining a practical model of speech recognition often requires a large amount of data for the learning (hundreds of hours, if not tens of thousands of hours). While in many scenarios it is often difficult and expensive to get enough audio training data, such as dialects. A straightforward solution is to use TTS for data augmentation. There has been a lot of research on TTS data augmentation methods for ASR tasks, such as \citep{laptev2020you,rossenbach2020generating,sun2020generating}.
In this work, we propose a unified framework for boosting low resource ASR tasks named MAC. The key in the MAC framework is leveraging a novel concatenative synthesis TTS system. It is worth mentioning that compared with former TTS for ASR tasks architectures, our MAC framework has the following advantages: 
\begin{itemize} 

\item  MAC leverages  a novel concatenative synthesis text-to-speech system to boost the ASR task, which can integrate language pronunciation rules as the prior knowledge. Furthermore, the process of generating audio has strong interpretability and is easy to be adjusted in the generation process; 
\item  Under the MAC framework, we propose a broad notion of meta audio set, and its dimensions can be flexibly determined according to prior knowledge, model complexity budget, audio data size, etc. So we can easily adjust the complexity and performance of the concatenative synthesis text-to-speech system.
\item  There is no need to train additional TTS neural networks, and the generation process is just simple splicing, which is easy to implement and saves computing resources; 
\item  Most importantly, MAC framework can be carried out in extremely low resource environments (e.g., training data less than 10h) without the help of additional labeled data. And MAC has a promising potential to model any low resource language as long as there is prior knowledge of the pronunciation rules of the language.
\end{itemize}

Lots of experiments have demonstrated the great effectiveness of our method in low resource ASR tasks. For the Cantonese ASR task, Taiwanese ASR task, and Japanese ASR task, the MAC framework can reduce the CER by more than 15\%.
Furthermore, MAC outperformed fine-tuning wav2vec2  on the Cantonese common speech dataset and obtained very competitive results on the Taiwanese and Japanese common speech datasets. Among them, we achieved a \textbf{10.9\%} character error rate (CER) on the common spoken Cantonese ASR task, resulting in a significant relative improvement of about \textbf{30\%} compared to fine-tuning wav2vec2 model.

Also, semi-supervised learning and transfer learning  are currently two main ways to deal with the (labeled) data limit problem. However, both of the  approaches will possibly face certain difficulties in some low resource scenarios. We will briefly discuss these below.

\subsection{Semi-supervised learning}

For semi-supervised learning, pseudo-label  algorithms are widely used. There are many variants of  pseudo label algorithms. For example, iterative pseudo label algorithm \citep{xu2020iterative}. But their basic ideas are using the pseudo labels to help to train the model.

Table \ref{table1} reported in \citet{higuchi2022momentum} demonstrated the  effectiveness of iterative pseudo-label semi-supervised algorithms. The experiments are conducted on the LibriSpeech dataset \citet{panayotov2015librispeech} and TED3 dataset \citet{hernandez2018ted}.
LL-10h, LS-100h, LS-360h and LS-860h in Table \ref{table1} represent the different splits of the LibriSpeech dataset \citet{panayotov2015librispeech} and the results are evaluated on dev-other split and test-other split of LibriSpeech dataset.

\begin{table}[ht]
\caption{Results of the iterative pseudo label based semi-supervised learning for ASR (reported in \citep{higuchi2022momentum}). The experiments are conducted on the LibriSpeech dataset \citet{panayotov2015librispeech} and TED3 dataset \citet{hernandez2018ted} with a character error rate (CER) evaluation criteria.\label{table1}\\}
\begin{center}

\begin{tabular}{ccc}

Resource & Dev  & Test \\
\hline
LS-100h               & 22.5 & 23.3 \\
LS-100h/LS-360h       & 15.9 & 15.8 \\
LS-100h/LS-860h       & \textbf{13.9} & \textbf{14.2} \\
\hline
LL-10h                & 50.6 & 51.3 \\
LL-10h/LS-360h        & 35.4 & 36.1 \\
LL-10h/LS-860h        & \textbf{33.5} & \textbf{34.4} 
\end{tabular}
\end{center}

\end{table}

Besides the fact that the performance can be indeed improved greatly, we also notice that the quality of the initial model performance (the first row in Table \ref{table1}) has a remarkable impact on the final performance. There are also some theoretical results to explain these phenomena (see e.g., \citep{wei2020theoretical} and \citep{min2022pseudo}). 
Actually, in real low resource scenarios, it can be rather difficult to obtain a suitable initial model (pseudo-label generator) due to the limited labeled  data. 
% Besides, there may not have enough unlabeled audio. 
These difficulties seriously affect the performance of pseudo label based semi-supervised learning when applied to extremely low resource ASR tasks.

\subsection{Transfer learning}

For fine-tuning pretrained large models, wav2vec2 \citep{baevski2020wav2vec,conneau2020unsupervised} is one of the most representative pretrained models.
Wav2vec2 has demonstrated strong transfer learning capabilities on ASR tasks. Using wav2vec2, the CER can be significantly reduced \citep{yi2021transfer}. Although transfer learning is generally quite effective, the effectiveness of transfer learning can be significantly constrained, if there is a large gap between the target speech domain and the pretrained speech domain. For example, the wav2vec2 \citet{baevski2020wav2vec} model is pretrained on English audio data in the English speech domain, which gives a 4.8\% CER by using 10 minutes of English audio labeled data to fine-tune. However, it only achieves a 28.32\% CER even using 27k training utts in the Japanese domain \citep{yi2021transfer}. 
 
\subsection{Our contributions}

The contributions can be summarized as follows:
\begin{itemize}
    \item We propose MAC framework, which leverages a novel concatenative synthesis text-to-speech system to boost the ASR task. And it can integrate language pronunciation rules as prior knowledge. 
    
    \item We innovatively proposed a broad notion of meta audio set, which helps MAC to integrate different prior knowledge and flexibly apply it to different scenarios.
    
    \item Mathematically, we give a clear description of MAC from the perspective of bayesian sampling. 
    
    \item Our experiments show that MAC brings more than 15\% reduction in CER in Cantonese ASR task, Taiwanese ASR task, and Japanese ASR task.
\end{itemize}

\section{Relate work}

There have been many attempts to use TTS for data augmentation to benefit ASR e.g. \citep{laptev2020you,rossenbach2020generating,sun2020generating,li2018training,ueno2021data,rosenberg2019speech,tjandra2017listening}.
Among them, \citep{ueno2021data} keep their eyes on the perspective of the representation. 
Results in \citet{rosenberg2019speech} indicate the effectiveness of TTS data enhancement, although it may not be as good as the model trained on real speech data.
\citep{tjandra2017listening} takes advantage of the close connection between TTS and ASR models.

However, current methods often suffer from severe difficulties in real speech recognition tasks in extremely low resource environments. For example, requiring additional data to train the TTS system or extremely low resource environments (around 10h) failing to meet the required audio of the models. In addition, the current NN based TTS systems often lack interpretability and are difficult to perform the necessary manual adjustment.

Recently, \citep{min202210} proposed a new interesting discovery. Experiments in \citet{min202210} show that competitive performance on Mandarin ASR tasks can be achieved with only 10h of Mandarin audio data with the following steps:
\begin{itemize}
    \item Train models on one Mandarin audio dataset and perform force alignment to get Mandarin character-audio pairs.
    \item Map the character to pinyin using character-pinyin dictionary to build pinyin-audio database.
    \item For transcription in another Mandarin audio dataset, query the audio clips corresponding to each Mandarin character and concatenate them.
    \item Perform energy normalization.
\end{itemize}
Although the \citet{min202210} is focused on Mandarin audio ASR tasks, it does have many properties that are well-suited for low resource ASR tasks. For example, it does not require other labeled audio data to work, which is valuable in the low resource environment. Besides the synthesis process is efficient, interpretable, and convenient for human intervention.

Inspired by the \citep{min202210}, we propose the MAC, leveraging a novel concatenative synthesis text-to-speech
system to boost the ASR task performance. In \citep{min202210}, the meta audio refers specifically to pinyin since only focuses on Mandarin ASR tasks. In this work, we propose a much more general concept of meta audio than \citep{min202210}. By specific meta audio, we can  utilize the prior knowledge of pronunciation rules in a large number of language scenes.
These advantages make MAC ideal for extremely low resource language ASR tasks. Furthermore, the work in  \citep{min202210}, focusing on Mandarin, can be regarded as a special case under the MAC framework. 
Extensive experiments prove the great effectiveness of MAC, particularly on low resource ASR tasks.

\section{Method}

Our basic idea is to generate the audio data by concatenating the meta audios corresponding to given transcription texts and then train the ASR neural network using the generated audio data. In our method, we propose a much more general concept of meta audio set. Based on the meta audio set conception
, our MAC can be expressed mathematically rigorously from the perspective of bayesian sampling.

\subsection{General audio datasets}
\label{sec3.1}

The mathematical formulation is as follows. Denote the audio wave space by $\mathcal{X}$, and the transcription text space by $\mathcal{Y}$, i.e., $\mathcal{X}=\{x:$all the audio waves$\},$ $\mathcal{Y}=\{y:$ all the transcription texts$\}.$

In general, for ASR tasks, the labeled audio dataset consists of audio-transcription pairs $\{(x_i,y_i)\}_{i=1}^N$ sampled from a certain underlying distribution $P$, and $\{x_i\}_{i=1}^N \sim P_x$, $\{y_i\}_{i=1}^N \sim P_y$ with $P_x,\, P_y$ as the marginal distribution of $P$. 

Obtaining a practical speech recognition model often requires hundreds even thousands of hours of audio-transcription pairs for training. Unfortunately, getting audio-transcription pairs is usually expensive. And in many scenarios such as dialects, one can only access around ten hours of audio-transcription pairs. However, the audio-only data $x \sim P_x$ and text-only data $y \sim P_y$ are respectively easier to access. Generally, we have a paired dataset $\mathcal{D}=\{(x_i,y_i)\}_{i=1}^N$, an audio-only dataset $\mathcal{D}_{\text{audio}}=\{x_i\}_{i=1}^{N_1}$ and a text-only dataset $\mathcal{D}_{\text{text}}=\{y_i\}_{i=1}^{N_2}$ with  $N_2 > N_1 > N$. This is the typical setting of low recourse ASR tasks. Now our goal is to sample paired audio-transcription data $(x,y)$ from the underlying distribution $P$.

Basically, we have 
\begin{equation}
    P(x,y)=P_y(y)P(x \,|\, y),
\end{equation}
where $P_y(y)$ corresponds to the distribution of transcriptions, and $P(x \,|\, y)$ denotes the distribution of audios given a certain transcription $y$. Therefore, the sampling of new data (audio-transcription pair) $(x,y)$ can be divided into two stages. First, the transcription text $y$ is sampled from $P_y(y)$, and then the audio $x$ corresponding to the previous transcription text $y$ is sampled from $P(x \,|\, y)$. The first stage, i.e., sampling of transcription text $y$, is relatively easy since usually there is sufficient text-only data in $\mathcal{D}_{\text{text}} = \{y_i\}_{i=1}^{N_2}$, and we can model/approximate $P_y$ by $\Tilde{P}_y(y)= \frac{1}{N_2} \sum_{y_i \in \mathcal{D}_{\text{text}}} \delta(y_i)$ (i.e., substitute the probability by sample frequency). Therefore, the key is to analyze and maximize $P(x \,|\, y)$, which is in fact a TTS task.

\subsection{Meta audio sequence space $\mathcal{A}$}
\label{sec3.2}
\begin{figure*}[ht]
\begin{center}
\includegraphics[width=\textwidth]{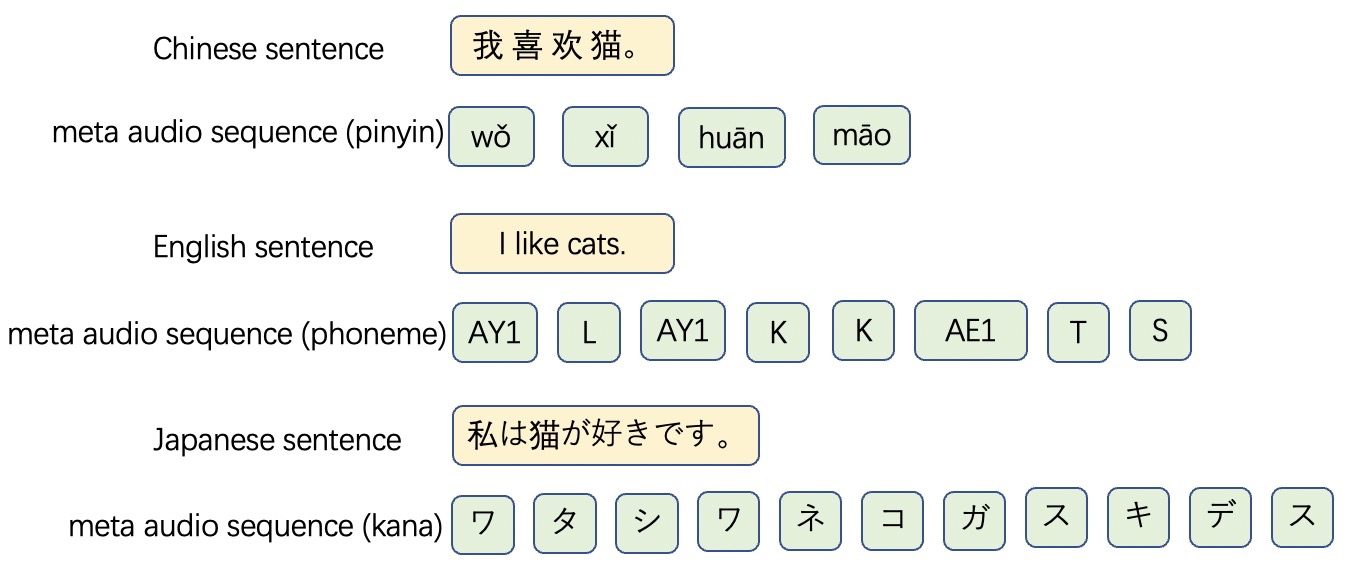}
\end{center}
\caption{Examples of the mapping $\mathbf{t}$ in English, Japanese and Chinese. The function $\mathbf{t}: \mathcal{Y} \to \mathcal{A}$ is to map a transcription text to its corresponding meta audio sequence which reflects language-specific pronunciation rules. Hence the  $\mathbf{t}$ can help to fusion prior knowledge of pronunciation rules. Obviously, different languages may have different mappings $\mathbf{t}$.
% transcription texts to corresponding meta audio sequences 
}
\label{fig1}
\end{figure*}

In order to perform further "decoupling" analysis on the conditional probability distribution $P(x \,|\, y)$, we first introduce the meta audio set. Here the meta audios  refer to the basic modeling units  of specific language pronunciations. For example, there are about 50 phonemes

in English. If we want to use these  phonemes as modeling units to characterize English pronunciation, these  phonemes form a nature meta audio set of English. In this way, it is similar to monophonic modeling in HMM-GMM \citet{rodriguez1997speech} and Kaldi \citet{povey2011kaldi}. However, in our framework, the meta audio set selections are flexible. For instance, a fusing phoneme (treating several  phonemes as the same phoneme) set can be seen as a meta audio set in English. For a specific language, we can reasonably determine the meta audio according to its pronunciation rules e.g. pinyin in Mandarin and kana in Japanese.

The meta audio sequence space $\mathcal{A}$ represents  sequences set of meta audios. Certainly, different meta audio sets may imply different meta audio sequence space $\mathcal{A}$. 

\paragraph{\textbf{Mapping function $\mathbf{t}: \mathcal{Y} \to \mathcal{A}$.}}
To reflect language-specific pronunciation rules, we also need a function $\mathbf{t}: \mathcal{Y} \to \mathcal{A}$ to map a transcription text to its corresponding meta audio sequence.
The construction of $\mathbf{t}$ requires prior knowledge of pronunciation rules. 
Obviously, different languages may have different mappings $\mathbf{t}$ and even different meta audio sequences $\mathcal{A}$ in the same language may have different mappings $\mathbf{t}$.  Figure \ref{fig1} shows examples of $\mathbf{t}$ for English and Japanese.

\subsection{Decoupling analysis}
\label{sec3.3}
\begin{figure*}[t!]
\begin{center}
\includegraphics[width=\textwidth]{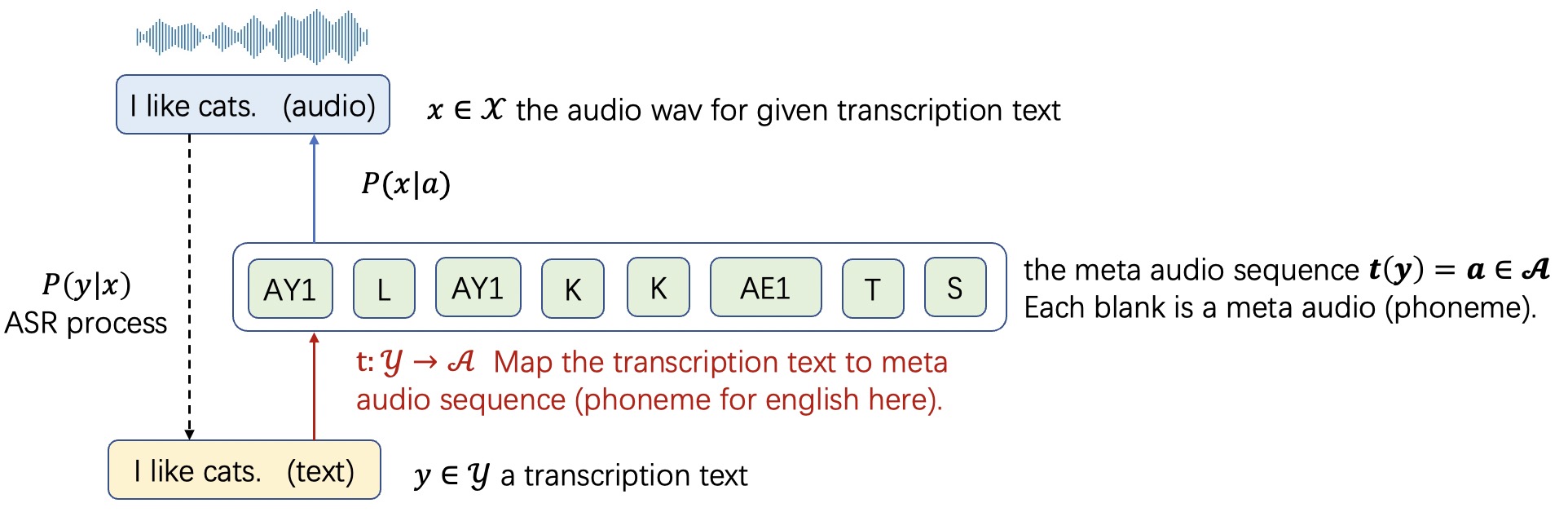}
\end{center}
\caption{ An illustration of the decoupling process. The goal of ASR task is to estimate $P(y \,|\, x)$ and the goal of TTS task is to estimate $P(x \,|\, y)$. In order to simplify the modeling of $P(x \,|\, y)$, we introduce a mapping $\mathbf{t}$ to transform $P(x \,|\, y)$ into $P(x \,|\, a)$. The dimension of $a$ is generally much smaller than $y$. }
\label{fig2}
\end{figure*}

In this section, we perform a fine-grained decoupling analysis on the probability $P(x\,|\,y)$, the conditional distribution of audios given transcriptions.
%and hence $P(x, y)$, the underlying distribution of audio-transcription pairs. 
On the one hand, when an audio $x \in \mathcal{X}$ is given, we have a conditional distribution $P(y \,|\, x)$. This is in fact the goal of ASR tasks: predict corresponding texts given audios by estimating $P(y \,|\, x)$. 
On the other hand, when a transcription $y \in \mathcal{Y}$ is given, we have another conditional distribution $P(x \,|\, y)$. We can decouple $P(x \,|\, y)$ via the meta audio sequence space $\mathcal{A}$ and the mapping function $\mathbf{t}$: 
\begin{equation}
\begin{aligned}
P(x \,|\, y)= & \sum_{a \in \mathcal{A}} P(x, a \,|\, y) \\
= & \sum_{a \in \mathcal{A}} P(x \,|\, a, y) P(a \,|\, y) \\
= & P(x \,|\, a=\mathbf{t}(y), y)
\end{aligned}
\end{equation}
Here, we applied the fact that $P(a \,|\, y)$ is a degenerate distribution with the probability 1 at $a=\mathbf{t}(y)$. Furthermore, the meta audio sequence contains all the pronunciation information of the transcription text, hence
\begin{equation}
    P(x \,|\, a=\mathbf{t}(y), y)=P(x \,|\, a=\mathbf{t}(y)),
\end{equation}
which gives
\begin{equation}\label{eq:Px|y=Px|a}
    P(x \,|\, y) = P(x \,|\, a=\mathbf{t}(y)).
\end{equation}
Figure \ref{fig2} illustrates this decoupling analysis.

\subsection{Bayesian inference on $P(x \,|\, a)$}
\label{sec3.4}
Based on Section \ref{sec3.3} and Eq. (\ref{eq:Px|y=Px|a}), 
instead of analyzing $P(x \,|\, y)$ directly, we can turn to study $P(x \,|\, a)$.
In general, this is much easier, since $a \in \mathcal{A}$ usually has a much lower dimension than $y \in \mathcal{Y}$. The dimension here refers to the number of element classes per position of $a \in \mathcal{A}$ or $y \in \mathcal{Y}$. For example, in English, the dimension of $a \in \mathcal{A}$ is about 50 (here we naturally select the phoneme as meta audio for simple interpretation), while the dimension of $y \in \mathcal{Y}$ may be much higher since there is a large number of words.

Notice that
\begin{equation}
    P(x \,|\, a)  \propto P_x(x) P(a \,|\, x),
\end{equation}
The goal is now converted to maximize $P_x(x)$ and $P(a \,|\, x)$ in order to maximize $P(x \,|\, a)$. Recall that $P_x(x)$ reflects the prior probability of audios $x \in \mathcal{X}$, we will discuss it later (in Section \ref{sec3.7}). For $P(a=(a^{(1)},a^{(2)},...,a^{(n)}) \,|\, x)$, we have
\begin{equation}
    \label{eq7}
    P(a \,|\, x) = \sum_{\mathbf{s}} \prod_{i=1}^n P\left(a^{(i)} \,|\, x^{(i)}=[x_{s_i}, x_{s_{i+1}})\right)
\end{equation}
Here, $\mathbf{s}=(s_1,s_2,...,s_{n+1})$ represents the time slice of $x \in \mathcal{X}$. Eq. (\ref{eq7}) holds because: 1) the audio wave $x$ can be also properly divided (maybe not unique) to obtain the clip $x^{(i)}=[x_{s_i}, x_{s_{i+1}})$ corresponding to each $a^{(i)}$. For instance, we can segment an English audio wave of one sentence and get the audio wave segmentation corresponding to the sentence's meta audio sequence (here meta audio sequence is phoneme sequence since here we naturally select the phoneme as meta audio for simple interpretation); 2) the audio wave clip $x^{(i)}$ in $x$ is monotonous with respect to $a^{(i)}$ in $a$. That is, the timestamps of audio waves and meta audios must match with each other, i.e., $x^{(i)}$ and only $x^{(i)}$ corresponds to $a^{(i)}$; 3) For simplicity, we treat this correspondence as independent.

Unfortunately, it is quite expensive to consider all the time slices $\mathbf{s}=(s_1,s_2,...,s_{n+1})$ in Eq. (\ref{eq7}). However, for each fixed time slice $\mathbf{s}^0=(s^0_1,s^0_2,...,s^0_{n+1})$, we can obtain a lower bound of $P(a \,|\, x)$:
\begin{equation}
\label{eq8}
\begin{aligned}
P(a \,|\, x) & = \sum_{\mathbf{s}} \prod_{i=1}^n P\left(a^{(i)} \,|\, x^{(i)}=[x_{s_i}, x_{s_{i+1}})\right) \\
 &\ge \prod_{i=1}^n P\left(a^{(i)} \,|\, x^{(i)}=[x_{s^{0}_i}, x_{s^{0}_{i+1}})\right).
 \end{aligned}
\end{equation}

\subsection{Maximization of $P(a \,|\, x)$}
\label{sec3.6}
\begin{figure*}[ht]
\begin{center}
\includegraphics[width=\textwidth]{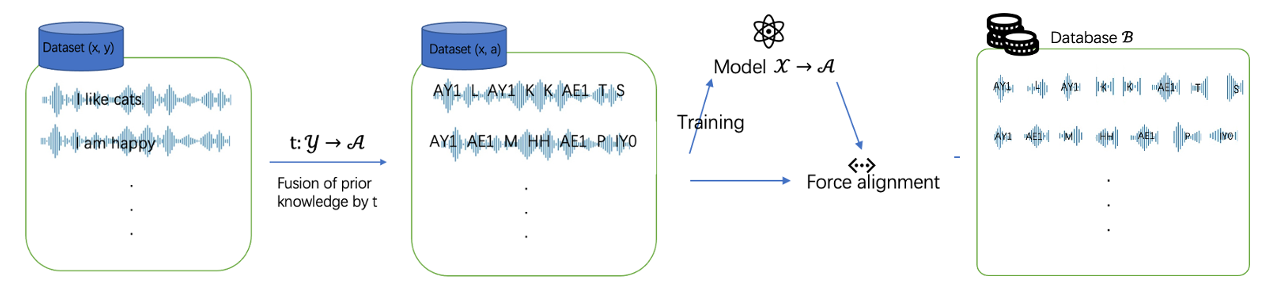}
\end{center}
\caption{ The process of building the database $\mathcal{B}$. Note that we can get audio clips from the training set, so we can train an ASR model on the training set and use that model for the forced alignment task on the (same) training set, which is very helpful in extremely low-resource conditions to obtain high-quality audio clips. }
\label{fig3}
\end{figure*}

According to Eq. (\ref{eq8}) in Section \ref{sec3.4}, we can approximately maximize $P(a=(a^{(1)},a^{(2)},...,a^{(n)})\,|\, x)$ by maximizing a lower bound determined by some fixed partition $\mathbf{s}^0=(s^0_1,s^0_2,...,s^0_{n+1})$. 
The maximization of right hand side of Eq. (\ref{eq8}) can be considered in another direction: for each element $a^{(i)}$ in $a$, we are required to find an audio wave clip $x^{(i)}$ that makes $P\left(a^{(i)} \,|\, x^{(i)}\right)$ as large as possible.
Fortunately, this can be achieved by performing force alignment \citep{kim2021reducing,lopez2022iterative}.
Specifically, we first map the transcription text $y \in \mathcal{Y}$ in the labeled audio dataset $\mathcal{D}=\{(x_i,y_i)\}_{i=1}^N$ into $a \in \mathcal{A}$ to get a corresponding dataset $\{(x_i,a_i)\}_{i=1}^N$. Then, we train the ASR model and perform force alignment on $\{(x_i,a_i)\}_{i=1}^N$ to get the audio wave clip corresponding to the meta audio element for each $a_i$ (with high probability). For further efficiency, we can store the forced alignment results and build a database $\mathcal{B}$. Therefore, when we aim to maximize $\prod_{i=1}^n P\left(a^{(i)} \,|\, x^{(i)}=[x_{s^{0}_i}, x_{s^{0}_{i+1}})\right)$ for some fixed time slice $\mathbf{s}$, we just query the audio clip $x^{(i)}$ corresponding to each meta audio element $a^{(i)}$. Here, the time slice $\mathbf{s}$ is considered implicitly, since when we concatenate these audio clips, it automatically forms a complete time of the audio wave. The procedure of building the database $\mathcal{B}$ is illustrated in Figure \ref{fig3}.

\begin{Remark}
\textbf{(Database size)} For each element $a^{(i)}$, we may get different corresponding audio clips (with high probability) by conducting force alignment on $\{(x_i,a_i)\}_{i=1}^N$. We just store all of them in the database $\mathcal{B}$ to enrich the selections and increase the diversity of synthesized audios.
\end{Remark}

\subsection{Maximization of $P_x(x)$}
\label{sec3.7}
In this section, we discuss about maximizing $P_x(x)$. In Section \ref{sec3.6}, for each element $a^{(i)}$ in $a$, we need to find an audio clip $x^{(i)}$ to enlarge $P\left(a^{(i)} \,|\, x^{(i)}\right)$ as much as possible. However, the audio wave obtained by combining these audio clips may give a higher probability $\prod_{i=1}^n P\left(a^{(i)} \,|\, x^{(i)} \right)$, but negatively affect $P_x(x=(x^{(1)},x^{(2)},...,x^{(n)}))$.
For example, the volume of generated $(x^{(1)},x^{(2)},...,x^{(n)})$ may change rapidly and frequently, leading to a low $P_x(x=(x^{(1)},x^{(2)},...,x^{(n)}))$. 
The reason can be mathematically understood that the support set of $P_x$ is likely to be a very small (proper) subset in the whole audio space $\mathcal{X}$. 
Hence, simply merging these audio clips may cause the synthesized audio wave $x=(x^{(1)},x^{(2)},...,x^{(n)})$ corresponding to $a=(a^{(1)},a^{(2)},...,a^{(n)})$ to be severely distorted.

To solve these problems, we need some regularization techniques to improve the modeling of $P_x(x=(x^{(1)},x^{(2)},...,x^{(n)}))$. 
A simple but effective regularization is the energy normalization \citep{li2002robust,ahadi2004energy,min202210}, and here we just imitate the  operation as in \citep{min202210}. That is, averaging the energy of sampled audio clips $(x^{(1)},x^{(2)},...,x^{(n)})$ corresponding to the meta audio sequence $a=(a^{(1)},a^{(2)},...,a^{(n)}) \in \mathcal{A}$. Mathematically, we have
\begin{equation}
    \label{eq9}
    E= \frac{\sum_{i=1}^{n}\lVert x^{(i)} \rVert}{n},
\end{equation}
\begin{equation}
\label{eq10}
    \left\{x^{(i)}\right\}_{i=1}^{n} \to \left\{\frac{x^{(i)}}{\lVert x^{(i)} \rVert}*E \right\}_{i=1}^{n}.
\end{equation}
Since the energy normalization in Eq. (\ref{eq10}) only linear scaling $x^{(i)}, i=1,2,...,n$, we further have
\begin{equation}
    P\left(a^{(i)} \,|\, x^{(i)}\right) \approx P\left(a^{(i)} \,|\, \frac{x^{(i)}}{\lVert x^{(i)} \rVert}*E \right).
\end{equation}

\begin{figure*}
    \centering
    \includegraphics[scale=0.65]{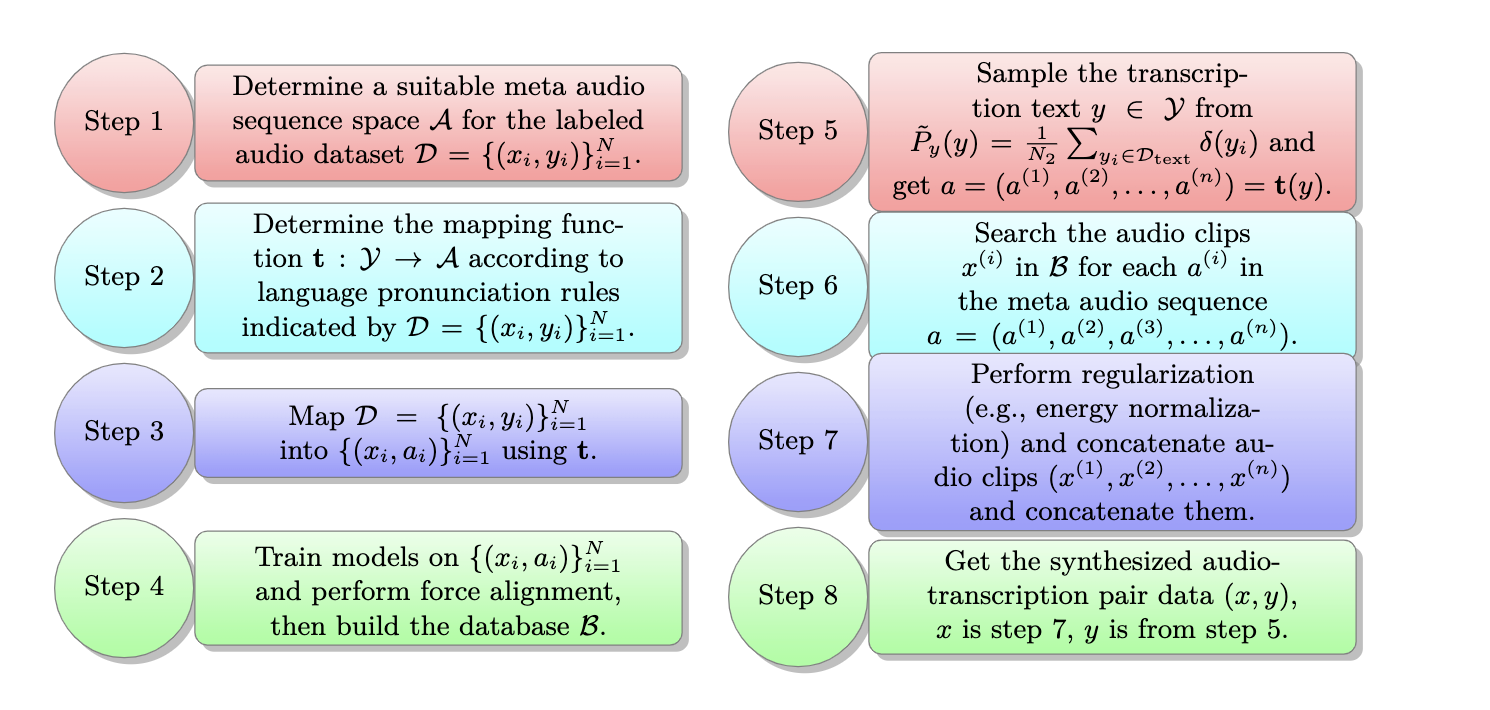}
    \caption{The MAC framework steps}
    \label{fig_step}
\end{figure*}

Certainly, other (more complex) regularization operations are also widely used in concatenative synthesis-based text-to-speech (TTS) systems \citep{tabet2011speech,khan2016concatenative}. In our experiments, the energy normalization is easy to implement and works well. 
More importantly, our ultimate goal is to do ASR tasks, so we do not pay too much attention to the fine-grained quality of the audio generated by TTS which may lead to overly complex models.
However, improving the quality of the synthesized audio may give better results and we will further discuss the applicability of other (more complex) regularization operations in the future work.

\subsection{Workflow of MAC}
\label{sec3.8}
According to previous sections, the MAC framework steps can be summarized as Fig \ref{fig_step}.

\section{Experiments}

We verify the effectiveness of MAC on three real low resource Cantonese, Taiwanese, and Japanese ASR tasks on the corresponding dataset of them in common voice dataset \footnote{We use the zh-HK dataset in the common voice dataset for the Cantonese ASR task. }. The experimental results show that MAC has defeated the results achieved by fine-tuning large-scale wav2vec2 pretrained models, and achieved very competitive results on these tasks. More importantly, MAC brings a remarkable performance boost on some tasks. For instance, we have achieved  \textbf{10.9\% CER} on the common voice dataset of Cantonese for ASR tasks, leading an around \textbf{30\% relative improvement} compared to fine-tuning wav2vec2 model.
\begin{figure*}
    \centering
    \includegraphics[scale=0.85]{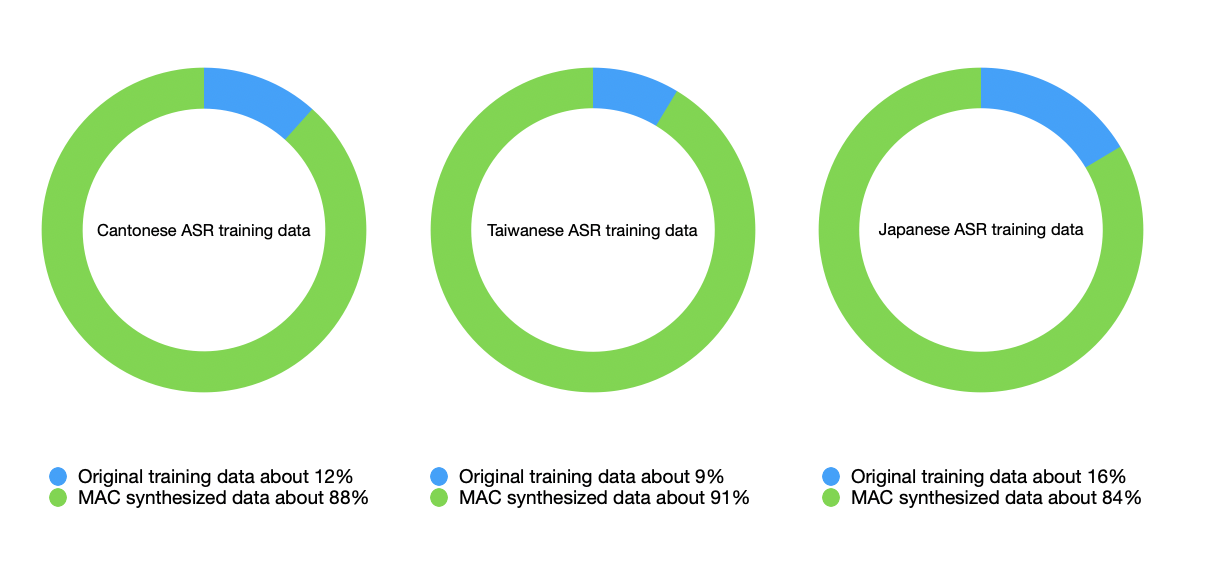}
    \caption{In the Cantonese ASR, Taiwanese ASR, and Japanese ASR three experimental tasks, the proportion of the audio data synthesized by MAC and the original audio data when we use MAC method.}
    \label{fig5}
\end{figure*}
\subsection{Datasets}

\begin{table}[bt]
\caption{CER for Cantonese, Taiwanese and Japanese on ASR tasks. We use the advanced hybrid CTC/attention model as the baseline (tested for four decoding modes: CTC greedy search, CTC prefix beam search, attention, and attention rescoring). It is shown that MAC can boost performance significantly, and also achieve competitive results compared to fine-tuning the pretrained wav2vec2 model.\\
\label{tab2}}
\centering
\begin{tabular}{c|c|c|c}
\hline
task                           & model                                            & decode mode            & CER   \\ \hline
\multirow{9}{*}{Cantonese ASR} & \multirow{4}{*}{Hybrid CTC/attention model}      & CTC greedy search      & 32.5 \\
                               &                                                  & CTC prefix beam search & 32.5 \\
                               &                                                  & attention              & 44.1 \\
                               &                                                  & attention rescoring    & 30.6 \\ \cline{2-4} 
                               & \multirow{4}{*}{Hybrid CTC/attention model+MAC} & CTC greedy search      & 12.7 \\
                               &                                                  & CTC prefix beam search & 12.7 \\
                               &                                                  & attention              & 11.0 \\
                               &                                                  & attention rescoring    & \textbf{10.9} \\ \cline{2-4}
                               & wav2vec2+fine-tuning                                & -                      & 15.4 \\ \hline
\multirow{9}{*}{Taiwanese ASR}   & \multirow{4}{*}{Hybrid CTC/attention model}      & CTC greedy search      & 51.3 \\
                               &                                                  & CTC prefix beam search & 51.3 \\
                               &                                                  & attention              & 55.3 \\
                               &                                                  & attention rescoring    & 48.2  \\ \cline{2-4} 
                               & \multirow{4}{*}{Hybrid CTC/attention model+MAC} & CTC greedy search      & 22.0 \\
                               &                                                  & CTC prefix beam search & 22.0 \\
                               &                                                  & attention              & 18.6 \\
                               &                                                  & attention rescoring    & 19.5 \\ \cline{2-4} 
                               & wav2vec2+fine-tuning                                & -                      & \textbf{18.4} \\ \hline
\multirow{9}{*}{Japanese ASR}    & \multirow{4}{*}{Hybrid CTC/attention model}      & CTC greedy search      & 45.3 \\
                               &                                                  & CTC prefix beam search & 45.3 \\
                               &                                                  & attention              & 72.1 \\
                               &                                                  & attention rescoring    & 44.5 \\ \cline{2-4} 
                               & \multirow{4}{*}{Hybrid CTC/attention model+MAC} & CTC greedy search      & 25.1 \\
                               &                                                  & CTC prefix beam search & 25.0  \\
                               &                                                  & attention              & 24.3 \\
                               &                                                  & attention rescoring    & \textbf{23.4} \\ \cline{2-4} 
                               & wav2vec2+fine-tuning                                & -                      & 24.9* \\ \hline
\end{tabular}

\end{table}
Common voice is currently one of the most widely used multilingual audio datasets, which is a public speech dataset contributed by volunteers around the world.
For each language, we use the training split of the corresponding dataset in common voice dataset as the origin labeled data to conduct MAC. The proportion of origin labeled data and synthetic data by MAC are shown in Fig. \ref{fig5}. For comparison, we also train directly on origin train split labeled data as the baseline results. We use the test split of the corresponding dataset in common voice dataset to evaluate.

\subsection{Models}

We use the advanced  hybrid CTC/Attention architecture  \citet{watanabe2017hybrid} with the conformer \citet{gulati2020conformer} encoder as Wenet \footnote{We refer to some codes from \url{https://github.com/wenet-e2e/wenet}} \citep{yao2021wenet}. 
It is worth mentioning that in the evaluation phase, the final output is jointly determined by both CTC-decoder and attention-decoder for better performance when we use the attention rescoring decode mode. That is, the attention decoder rescores the top candidates given by the CTC-decoder. We conduct experiments on 2 $\times$ RTX 3090GPUs (24GB) and 4 $\times$ P100GPUs (16GB).

\subsection{Details and results}

Below we present the experimental details and results

We can see that our MAC method substantially exceeds the fine-tuning wav2vec2 model result on the Cantonese recognition task and is comparable to the wav2vec2 model results on other tasks.

\subsubsection{Cantonese ASR}
\label{sec4.3}
For Cantonese, the natural choice of the meta-audio set is  Cantonese pinyin at first glance. But taking advantage of the broad notion of the meta audio set, we make some adjustments on the Cantonese pinyin to simplify our modeling process, including but not limited to not distinguishing between different tones in Cantonese pinyin in the meta audio set. e.g. regarding "haa4" and "haa6" as one meta audio in the meta audio set. The function $\mathbf{t}$ is to map a Cantonese transcription text $y$ to the meta audio sequence $a \in \mathcal{A}$. For a Cantonese transcription text $y=(y^1,y^2,...,y^m)$, we generally have the approximation
\begin{equation}
\label{eq12}
    \mathbf{t}(y^1,y^2,...,y^m) \approx (\mathbf{t}(y^1),\mathbf{t}(y^2),...,\mathbf{t}(y^m)).
\end{equation}
Hence, we can reverse the order of Step 3 and Step 4 in Section \ref{sec3.8} to simplify the overall process. That is, we first train the model on original labeled dataset $(x, y)$ and perform forced alignment. And then pass the results through the mapping function $\mathbf{t}$ to build the database $\mathcal{B}$ for Cantonese. In this way, we can use the trained baseline Hybrid CTC/attention model to do force alignment and do not need to train another model for force alignment. We use transcriptions in the validation split of the  zh-hK dataset in common voice dataset and remove transcriptions appearing in the test split as the text-only dataset  $\mathcal{D}_{\text{text}}$ for Step 5. For Step 7, we use the energy normalization method described in Eq. (\ref{eq9}) and (\ref{eq10}).

\subsubsection{Taiwanese ASR}
For Taiwanese, the natural choice of the meta-audio set is  pinyin at first glance. But taking advantage of the broad notion of the meta audio set, we make some adjustments on the Cantonese pinyin to simplify our modeling process, including but not limited to simplification of some polyphonic characters. The function $\mathbf{t}$ is to map a Taiwanese transcription text $y \in \mathcal{Y}$ to a meta audio sequence $a \in \mathcal{A}$. The pinyin mapping function $\mathbf{t}$ also has approximation Eq. (\ref{eq12}) and hence we can also reverse the order of Step 3 and Step 4 to simplify the overall process. 
% Other details are the same as those of Cantonese in Section \ref{sec4.3}.
That is as Cantonese task above, we first train the model on original labeled dataset $(x, y)$ and perform forced alignment, and then pass the results through the mapping function $\mathbf{t}$ to build the database $\mathcal{B}$ for Taiwanese. In this way, we can use the trained baseline Hybrid CTC/attention model to do force alignment and do not need to train another model for force alignment. We use transcriptions in the validation split of the taiwan dataset in common voice dataset and remove transcriptions appearing in the test split as the text-only dataset  $\mathcal{D}_{\text{text}}$ for Step 5. For Step 7, we use the energy normalization method described in Eq. (\ref{eq9}) and (\ref{eq10}).

\subsubsection{Japanese ASR}

For Japanese, the natural choice of the meta-audio set is kana at first glance. But taking advantage of the broad notion of the meta audio set, we make some adjustments on the Cantonese pinyin to simplify our modeling process, including but not limited to simplification of some special kanas. The function $\mathbf{t}$ is to map a Japanese transcription text $y \in \mathcal{Y}$ to a kana sequence $a \in \mathcal{A}$. Here, we do not reverse Step 3 and Step 4. 
% Other details are the same as those of Cantonese.
We use transcriptions in the validation split of the  japan dataset in common voice dataset and remove transcriptions appearing in the test split as the text-only dataset  $\mathcal{D}_{\text{text}}$ for Step 5. For Step 7, we use the energy normalization method described in Eq. (\ref{eq9}) and (\ref{eq10}).

\subsubsection{Performance}

All results are shown in Table \ref{tab2}. We just refer to fine-tuning wav2vec2 results from urls \footnote{ we refer the fine-tuning wav2vec2 results on urls. Cantonese: \url{https://huggingface.co/ctl/wav2vec2-large-xlsr-cantonese} and Japanese: \url{https://huggingface.co/qqhann/w2v_hf_jsut_xlsr53} and Taiwanese: \url{https://huggingface.co/voidful/wav2vec2-large-xlsr-53-tw-gpt} and we round to one decimal place.} so we did not apply the  decode modes (CTC greedy search, CTC prefix beam search, attention, attention rescoring) to them. Firstly, one can observe a significant performance improvement by MAC compared to the advanced hybrid CTC/attention baseline model. For \textbf{all} CTC greedy search, CTC prefix, attention, and attention rescoring decode mode  in \textbf{all} three Cantonese ASR task, Taiwanese ASR task, and Japanese ASR task, the MAC method can reduce the CER by more than 15\%.

More importantly, MAC beats wav2vec2 (with fine-tuning) and achieves new SOTA on common voice datasets of Cantonese and Japanese when applied to ASR tasks. * in Table \ref{tab2} represents that the 24.9 CER result is achieved by using \emph{extra data} than the Japanese audio dataset in the common voice dataset during  finetuning wav2wev2. We can see that MAC relatively improves the performance by about 30\% on the Cantonese ASR task and even beats wav2vec2 fine-tuned with extra data on the Japanese ASR task. In addition, for the Taiwanese ASR task, MAC also achieves results comparable to the fine-tuned wav2vec2 model.

\section{Conclusion}

MAC proposed in this work is a unified framework for low resource automatic speech recognition tasks.
We propose a broad notion of meta audio set to help MAC to be available as long as we have the knowledge of pronunciation rules to construct a suitable meta audio set.
Hence it can  meet the modeling needs of numerous languages and numerous scenes.
Besides,  we give a clear mathematical description of MAC framework from the perspective of bayesian sampling.
Our experiments have demonstrated the great effectiveness of MAC on low resource speech recognition tasks, with remarkable improvements in accuracy even without tuning hyperparameters carefully.
We hope that the MAC method can contribute to the development of low resource audio recognition tasks.

\bibliography{main}
\bibliographystyle{tmlr}

\end{document}